\documentclass{article}

\usepackage[preprint]{neurips_2026}
\makeatletter
\renewcommand{\@noticestring}{} 
\makeatother
\usepackage{booktabs}
\usepackage{amsmath}
\usepackage{tabularx}
\usepackage[utf8]{inputenc} 
\usepackage{subcaption}
\usepackage[T1]{fontenc}    
\usepackage{hyperref}       
\usepackage{url}            
\usepackage{booktabs}       
\usepackage{amsfonts}       
\usepackage{nicefrac}       
\usepackage{microtype}      
\usepackage{xcolor}         
\usepackage{listings}
\usepackage{multirow}          
\usepackage{xcolor}
\usepackage{float}
\usepackage{graphicx} 

\usepackage[utf8]{inputenc}

\newcommand\blfootnote[1]{%
  \begingroup
  \renewcommand\thefootnote{}\footnote{#1}%
  \addtocounter{footnote}{-1}%
  \endgroup
}

\title{RAFT: Data Refinement and Adaptive Distillation for Domain Fine-Tuning with Alleviated Forgetting}

\author{
Yuduo Li$^{1,2 * \mathsection}$,
Xiaofeng Shi$^{1,2 * \dagger}$,
Qian Kou$^{1}$,
Longbin Yu$^{1}$,
Hua Zhou$^{1 \ddagger}$ \\[0.5em]
$^{1}$Beijing Academy of Artificial Intelligence (BAAI) \\
$^{2}$Beijing Jiaotong University (BJTU)
}

\date{}
\begin{document}

\maketitle
\blfootnote{$^*$ Equal contribution.}
\blfootnote{$^\mathsection$ Work done during internship at BAAI.}
\blfootnote{$^\dagger$ Corresponding author. Email: \texttt{xfshi@baai.ac.cn}}
\blfootnote{$^\ddagger$ Project leader.}


\begin{abstract}
Domain-specific supervised fine-tuning (SFT) often improves in-domain performance at the cost of degrading a model's general capabilities. We view this degradation through two practical gaps in domain SFT: a supervision-compatibility gap, where domain targets differ in style and reasoning format from the original model's natural responses, and a trajectory-preservation gap, where teacher-forced SFT optimizes fixed target tokens without constraining the model's behavior on its own generated prefixes. This process fails to preserve the model's original behavior. We propose RAFT (Data \textbf{R}efinement and \textbf{A}daptive Distillation for Domain \textbf{F}ine-\textbf{T}uning with Alleviated Forgetting), a two-stage framework that addresses both factors. First, RAFT constructs model-compatible supervision through self-conditioned rewriting, semantic filtering, and answer fusion. Second, RAFT performs Answer-Conditioned On-Policy Distillation, where the original instruction-tuned model provides soft targets on student-generated trajectories while being conditioned on the fused answer as helpful context. We further introduce top-K temperature distillation and EMA-based adaptive loss balancing to stabilize the domain-general trade-off. Across three instruction-tuned backbones and five domains, RAFT improves average domain accuracy by 23.2\% over standard SFT, while recovering part of the SFT-induced degradation on MS-Bench and IFEval, with relative improvements of 18.2\% and 10.2\%, respectively. These results show that coupling data refinement with trajectory-level preservation provides an effective recipe for domain fine-tuning with alleviated forgetting.

\end{abstract}

\section{Introduction}
\label{sec:intro}
Large language models (LLMs) acquire broad capabilities through pre-training and instruction tuning~ \citep{zhang2026instruction,grattafiori2024llama,abouelenin2025phi}, yet domain-specific supervised fine-tuning (SFT) often causes \emph{catastrophic forgetting}~ \citep{luo2025empirical,kotha2023understanding}---general capabilities degrade substantially. We focus on two practical factors that make domain SFT prone to forgetting: (1)~\emph{distribution mismatch} between ground-truth supervision and the model's own output distribution, forcing abrupt parameter shifts that overwrite pre-trained knowledge~ \citep{yang2024self}; and (2)~the absence of explicit mechanism to \emph{preserve general capabilities} during optimization, which focuses exclusively on domain loss.

Many existing methods primarily emphasize one side of this problem. Self-distillation (SDFT)~ \citep{yang2024self} rewrites supervision to reduce distribution mismatch but lacks training-level regularization. On-policy distillation  \citep{lu2025onpolicydistillation} introduces KL constraints on student trajectories, but heavily relies on the teacher model's quality, making it sensitive to teacher-bias. Replay-based methods~ \citep{lu2026mssr,luo2025empirical} mix general-domain data but are sensitive to mixing ratios. Regularization approaches such as L2SP~ \citep{xuhong2018explicit} and EWC~ \citep{kirkpatrick2017overcoming} add parameter-level constraints that ignore functional behavior and are costly for LLMs. Forgetting-aware pruning~ \citep{huang2025mitigating} provides structural regularization alone. In short, existing methods usually focus on either data quality or training constraints, but rarely both. While self-distillation and KL divergence are well-established, how to effectively combine data refinement with training-level regularization remains a gap in domain SFT.

The key challenge is that these two aspects are coupled rather than independent. Data refinement improves the reference answers used for cross-entropy training, but does not constrain the model's behavior on its own generated trajectories. Conversely, on-policy distillation regularizes student trajectories, but without target-aware supervision it can act as a conservative anchor that limits domain adaptation. RAFT addresses this coupling by using the fused answer both as model-compatible SFT supervision and as helpful context for the teacher during on-policy distillation.

\begin{figure*}[t!]
    \centering
    \includegraphics[width=\textwidth]{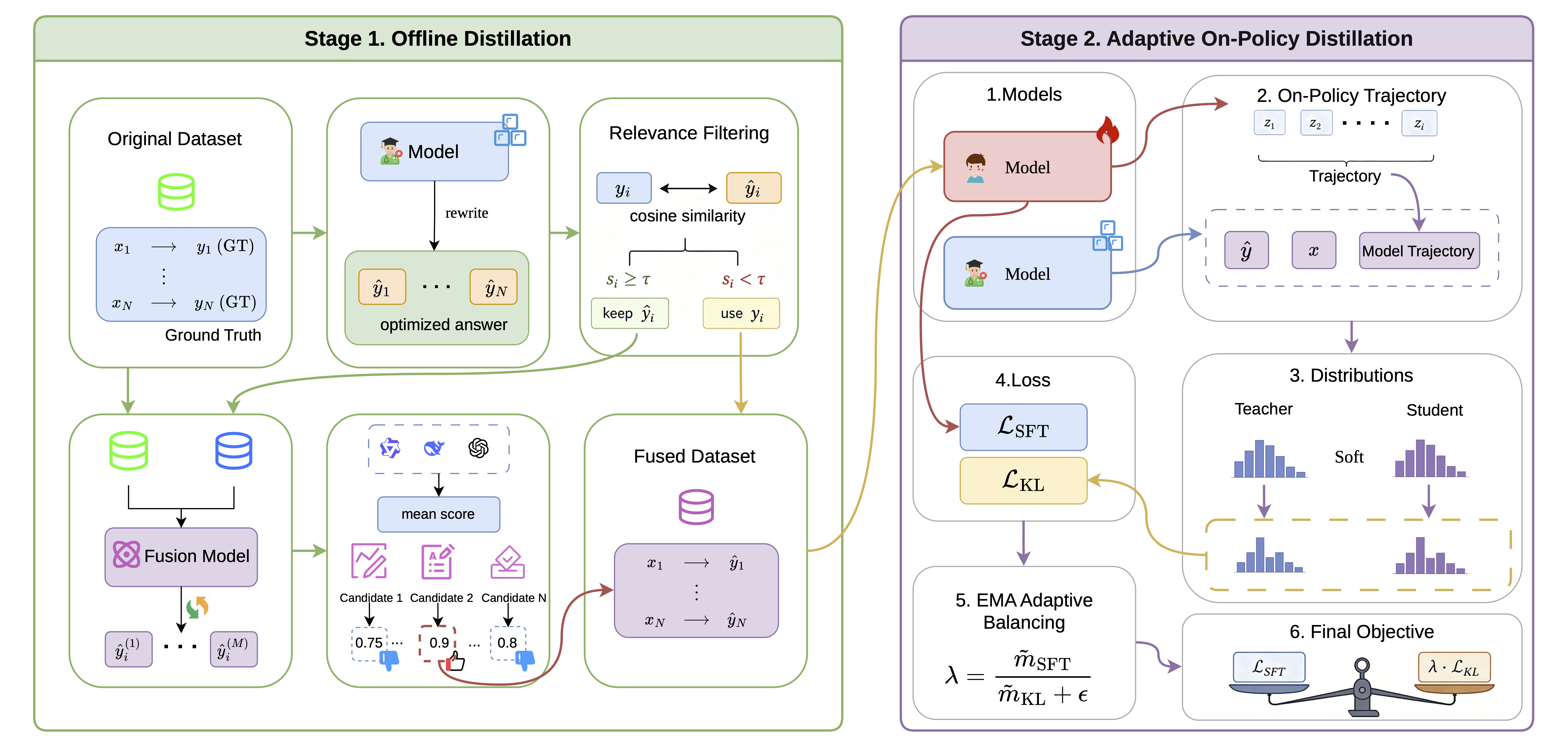}
    \vspace{-3pt}
    \caption{Overview of the \textbf{RAFT} framework. \textbf{Left: Offline Distillation} generates higher-quality fused data by combining the model's rewritten answers (filtered by cosine similarity) with the original data through a stronger fusion model. \textbf{Right: Adaptive on-policy Distillation} trains the model with Answer-Conditioned On-Policy Distillation, distillation over softened probability distributions, and EMA-based adaptive loss balancing.}
    \label{fig:raft}
\end{figure*}

As illustrated in Figure~\ref{fig:raft}, RAFT combines data optimization and training optimization through offline distillation and adaptive on-policy distillation. In \emph{offline distillation}, we first rewrite the ground-truth answers conditioned on the model's own distribution, and use cosine-similarity filtering to determine whether to adopt the rewritten answers. The selected data is combined with the original samples and input into a stronger fusion model to generate refined responses. In \emph{adaptive on-policy distillation}, we use the fused dataset to train the model with explicit general capability preservation. At each training step, the model first generates trajectories via autoregressive sampling. The original instruction-tuned model (teacher) then provides soft supervision through Answer-Conditioned On-Policy Distillation, where it is conditioned on the answer in the Fused Dataset as additional context for more informative guidance. We further introduce a top-$K$ temperature distillation mechanism to focus KL divergence on the most informative tokens, which prevents over-smoothing and improves output diversity. In addition, we propose an EMA-based adaptive balancing strategy that dynamically adjusts the SFT--distillation weight, eliminating the need for manual hyperparameter tuning.


We evaluate RAFT on three models (SmolLM3-3B, Llama-3.2-3B-Instruct, Phi-4-mini-instruct) across five domains. Our contributions are:
\begin{itemize}
\item We identify and target two practical factors associated with forgetting in domain SFT: supervision distribution mismatch and the lack of explicit preservation constraints.

\item We introduce an offline distillation stage that uses cosine similarity filtering to select distribution-consistent rewritten answers and fuses them with original data to improve response quality.

\item We propose an adaptive on-policy distillation method that combines Answer-Conditioned On-Policy Distillation, top-$K$ temperature distillation mechanism, and EMA-based loss balancing.

\item RAFT recovers part of SFT-induced degradation, as measured by MS-Bench (18.2\%) and IFEval (10.2\%), while also improving domain accuracy (D-Acc) by 23.2\%.
\end{itemize}

\section{Related Work}
\paragraph{Mitigating forgetting of general capabilities.}
Fine-tuning LLMs for domain-specific tasks often leads to catastrophic forgetting of their general capabilities  \citep{qi2023fine}. Existing mitigation methods can be broadly grouped into three categories. (1) replay-based methods  \citep{rolnick2019experience,huang2024mitigating} alleviate forgetting by mixing general-domain data or replay samples with task-specific training data, so that the model continues to rehearse its pretrained knowledge during adaptation. (2)  regularization-based methods  \citep{chen2020recall,zixuan2023continual} reduce destructive drift from the pretrained model by constraining parameter updates or output distributions. (3) parameter-space adjustment methods restrict adaptation to a limited subset of parameters or carefully modify model weights, for example through parameter-efficient tuning such as LoRA  \citep{hu2022lora}, partial parameter freezing  \citep{huang2025mitigating}, thereby preserving more of the model's original general abilities.

\paragraph{On-policy distillation.}
Online distillation aligns the student with teacher signals on trajectories generated by the current student policy, rather than relying on a fixed offline dataset \citep{shenfeld2026selfdistillationenablescontinuallearning}. By matching supervision to the model's own trajectories during training, it has shown strong potential for preserving general capabilities  \citep{hubotter2026reinforcement,lu2025onpolicydistillation}. However, pure online distillation does not fully leverage the labeled target answers provided in standard supervised datasets \citep{agarwal2023gkd}. In contrast, our approach combines SFT supervision with on-policy distillation, allowing the model to learn domain-specific answers accurately while alleviating catastrophic forgetting.A conceptual comparison with these strategies is provided in Appendix~\ref{app:method-comparison}.

\section{Methodology}

\subsection{Overview}
\label{sec:overview}

We propose a unified framework for mitigating catastrophic forgetting in domain-specific supervised fine-tuning (SFT). As discussed in Section~\ref{sec:intro}, catastrophic forgetting arises from both data stage and optimization stage causes. Accordingly, our framework intervenes at two stages: In the data stage, constructing higher-quality training data through offline distillation (Section~\ref{sec:fusion}); In the optimization stage, adding distillation constraints through adaptive on-policy distillation (Section~\ref{sec:distillation}).

Given a domain dataset $\mathcal{D}=\{(x_i, y_i)\}_{i=1}^{N}$, where $x_i$ is the input instruction and $y_i$ is the ground truth answer. Let $\theta$ denote the parameters of the model being fine-tuned. Standard SFT optimizes $\theta$ only on domain ground truth answers, which improves domain performance but causes the model to deviate from the knowledge and behavioral patterns acquired during pre-training and alignment.

\subsection{Offline Distillation}
\label{sec:fusion}

A direct cause of catastrophic forgetting is the distribution mismatch between the ground truth answers and the model's output distribution before fine-tuning: forcing the model to fit answers far from its natural expression style leads to abrupt parameter shifts that overwrite pre-trained knowledge. To address this, we construct training data that preserves factual content while adapting the expression style to the original instruction-tuned model, proceeding through three stages: self-conditioned rewriting, semantic filtering, and multi-expert optimal selection over fused responses.

\textbf{Self-Conditioned Rewriting.} For each sample $(x_i, y_i)$, the original instruction-tuned model generates a rewrite conditioned on the original answer:
\begin{equation}
\tilde{y}_i \sim p_{\theta}(\cdot \mid x_i, y_i)
\end{equation}
where $p_{\theta}$ denotes the output distribution of the original instruction-tuned model parameterized by $\theta$. Conditioning on $y_i$ preserves key factual content while adapting the expression style to that of the original instruction-tuned model's own output distribution, following the self-distillation paradigm~ \citep{yang2024self}. However, self-rewriting without quality control can introduce semantic drift, where the rewritten answer deviates from the original meaning.

 \textbf{Semantic Filtering.} To prevent semantic drift from self-rewriting, we compute the cosine similarity between the original
   answer and the rewritten answer in a shared embedding space:                                                                    
  \begin{equation}                                                                                                                 
  s_i = \frac{\phi(y_i)^\top \phi(\tilde{y}_i)}{\|\phi(y_i)\| \cdot \|\phi(\tilde{y}_i)\|}                                         
  \end{equation}                                                                                                                   
  where $\phi: \mathcal{V}^* \to \mathbb{R}^d$ is a pre-trained sentence encoder that maps text sequences to $d$-dimensional       
  embeddings. Candidate rewrites are filtered through a threshold:                                                                 
  \begin{equation}                                                                                                                 
 \bar{y}_i =                                 \begin{cases}                                                           
  \tilde{y}_i, & s_i \ge \tau \\                                                     
  y_i, & s_i < \tau                                                             
  \end{cases}                                                   
  \end{equation}                                                                                                                   
  where $\tau$ is the similarity threshold (see Appendix \ref{appendix:cos_threshold} for detailed analysis). When the rewrite is  
  semantically consistent with the original answer ($s_i \ge \tau$), $\bar{y}_i$ is retained for subsequent fusion with $y_i$;     
  otherwise, the original answer $y_i$ is used directly, preventing error accumulation in self-reinforcement. While semantic       
  filtering ensures per-sample quality, individual model outputs still exhibit variance due to the stochastic nature of            
  generation---a single fusion attempt may not yield the best possible result.

  \textbf{Multi-expert optimal selection over fused responses.} Considering the stochastic nature of model outputs, we perform multiple      
  rounds of fusion to improve robustness and output quality. Specifically, for samples that pass semantic filtering ($s_i \ge
  \tau$), we fuse $y_i$ and $\bar{y}_i$ through a fusion model (Qwen3-3B \citep{qwen3technicalreport}) $f_{\text{merge}}$ to        
  generate a higher-quality answer $\hat{y}_i$(see Appendix~\ref{app:fusion_prompt} for the prompt template). To ensure the quality of the fused output, we employ multiple judge models to
  independently evaluate the consistency between the fused result and the original answer. We perform multiple rounds of fusion and
   scoring, and select the result with the highest average evaluation score as the final $\hat{y}_i$. Details of the evaluation are
   provided in Appendix~\ref{appendix:Quality_Scoring}. Finally, we construct the dataset:
  \begin{equation}
  \hat{\mathcal{D}} = \{(x_i, \hat{y}_i)\}_{i: s_i \ge \tau} \cup \{(x_i, y_i)\}_{i: s_i < \tau}                                   
  \end{equation}                                                                                                                   
  which preserves Factual integrity while adapting the expression style to the instruction-tuned model's       
  distribution, ensuring output quality through the multi-model scoring mechanism.        
  
\subsection{Adaptive On-Policy Distillation}
\label{sec:distillation}

Although the fused data $\hat{\mathcal{D}}$ alleviates distribution mismatch in the offline distillation stage, SFT alone still suffers from catastrophic forgetting: optimizing solely on domain data causes the model to gradually lose its general capabilities. To mitigate this fundamental limitation, we adopt the original instruction-tuned model itself as the teacher and perform online distillation, ensuring that while the student model learns domain-specific knowledge, it is also constrained to preserve the behavioral patterns of the original model.

\textbf{SFT.} We optimize the standard cross-entropy loss on $\hat{\mathcal{D}}$. Let $\mathcal{A}_i$ denote the set of answer token positions in the $i$-th sample, and $\hat{y}_{i,t}$ denote the $t$-th token of the fused answer $\hat{y}_i$:
\begin{equation}
\mathcal{L}_{\text{SFT}} = -\frac{1}{N}\sum_{i=1}^{N}\frac{1}{|\mathcal{A}_i|}\sum_{t \in \mathcal{A}_i} \log p_\theta(\hat{y}_{i,t} \mid x_i, \hat{y}_{i,<t})
\end{equation}
This loss injects the fused domain knowledge into $\theta$. However, this loss alone provides no mechanism to preserve the student model's general capabilities. To address this, we introduce distillation constraints that constrain the student model's behavior on its own generated trajectories.

\textbf{Online Policy Trajectory Generation.} For each $x_i$, the student model generates a token sequence as a response trajectory through autoregressive sampling according to the current policy:
\begin{equation}
\mathbf{z}_i = (z_{i,1}, \ldots, z_{i,L_i}) \sim p_\theta(\cdot \mid x_i)
\end{equation}
where each $z_{i,t} \in \mathcal{V}$ is the token sampled by the student model at step $t$, and $L_i$ is the trajectory length. Unlike imposing constraints on reference answer trajectories, online policy distillation ensures that the constraints cover the state space that the student model would access during actual inference. Having generated the student's trajectories, we next need to define how the teacher provides supervision on these trajectories.

\textbf{Answer-Conditioned On-Policy Distillation.} Since the student's trajectories may deviate from what the original instruction-tuned teacher would naturally produce, the teacher model is given the fused answer as additional context so that it can provide more informative guidance at each step. The student model predicts the next token conditioned only on $x_i$, while the teacher model uses the fused answer as additional context. At position $t$ of trajectory $\mathbf{z}_i$:
\begin{equation}
p_\theta^S(\cdot \mid x_i, z_{i,<t}), \qquad p_\psi^T(\cdot \mid r, \hat{y}_i, x_i, z_{i,<t})
\end{equation}
where $r$ is the reference prompt, and the teacher model takes as input the sequential concatenation of $r$, $\hat{y}_i$, $x_i$, and $z_{i,<t}$. Both models predict on the same trajectory prefix $z_{i,<t}$, but the teacher can provide more informative soft targets at each step of the student's trajectory by conditioning on $\hat{y}_i$. See Appendix~\ref{appendix:template} for the teacher model's concatenation template. With the teacher providing soft targets on the student's own trajectories, we next define how the student's distribution is guided toward the teacher's distribution while maintaining output diversity.

\textbf{Top-K Temperature Distillation.} To increase the diversity of model outputs and prevent the model from concentrating all probability mass on a single token, we apply temperature scaling with top-$K$ selection to smooth the output distributions. Let $\mathbf{u}_{i,t}, \mathbf{v}_{i,t} \in \mathbb{R}^{|\mathcal{V}|}$ denote the logits of the teacher and student models at position $(i,t)$ respectively, and $T > 0$ be the temperature parameter. The temperature-scaled distributions over vocabulary $\mathcal{V}$ are:
\begin{equation}
q_{i,t}(a) = \frac{\exp(u_{i,t}(a)/T)}{\sum_{b \in \mathcal{V}}\exp(u_{i,t}(b)/T)}, \quad p_{i,t}^{(T)}(a) = \frac{\exp(v_{i,t}(a)/T)}{\sum_{b \in \mathcal{V}}\exp(v_{i,t}(b)/T)}
\end{equation}
We select the top-$K$ tokens with highest probability from the teacher distribution to form the set $\mathcal{K}_{i,t} = \text{TopK}(q_{i,t}, K)$, and restrict both distributions to this set with renormalization:
\begin{equation}
\tilde{q}_{i,t}(a) = \frac{q_{i,t}(a)\,\mathbf{1}[a \in \mathcal{K}_{i,t}]}{\sum_{b \in \mathcal{K}_{i,t}} q_{i,t}(b)}, \quad \tilde{p}_{i,t}(a) = \frac{p_{i,t}^{(T)}(a)\,\mathbf{1}[a \in \mathcal{K}_{i,t}]}{\sum_{b \in \mathcal{K}_{i,t}} p_{i,t}^{(T)}(b)}
\end{equation}
This focuses distillation on the $K$ tokens that the teacher considers most important while reducing the computational cost of full-vocabulary KL, and the temperature scaling ensures output diversity by smoothing the distribution rather than collapsing it to a single token. The online policy distillation loss is:
\begin{equation}
\mathcal{L}_{\text{KL}} = \frac{T^2}{N}\sum_{i=1}^{N}\frac{1}{L_i}\sum_{t=1}^{L_i} \text{KL}(\tilde{q}_{i,t} \| \tilde{p}_{i,t})
\end{equation}
where $T^2$ is the standard compensation factor for temperature scaling, used to compensate for the gradient magnitude reduction after temperature transformation. Having defined both the SFT loss and the distillation loss, we now face the challenge of balancing them: since $\mathcal{L}_{\text{SFT}}$ uses cross-entropy and $\mathcal{L}_{\text{KL}}$ uses KL divergence, their magnitudes differ in scale and vary dynamically during training.

\textbf{EMA-based Adaptive Balancing.} We want the two losses to remain stable and equally important throughout training, rather than having their relative contributions shift unpredictably as the data and model change. To this end, we introduce an exponential moving average (EMA)-based adaptive balancing mechanism. At training step $s$, with decay coefficient $\beta \in (0,1)$:
\begin{equation}
m_k^{(s)} = \beta\, m_k^{(s-1)} + (1-\beta)\,\mathcal{L}_k^{(s)}, \quad \hat{m}_k^{(s)} = \frac{m_k^{(s)}}{1-\beta^s}, \quad k \in \{\text{SFT}, \text{KL}\}
\end{equation}
where $m_k^{(s)}$ is the EMA estimate of the loss, and $\hat{m}_k^{(s)}$ is bias-corrected by dividing by $1-\beta^s$ to address early underestimation caused by zero initialization. The adaptive weight and final objective are:
\begin{equation}
\lambda^{(s)} = \frac{\hat{m}_{\text{SFT}}^{(s)}}{\hat{m}_{\text{KL}}^{(s)} + \epsilon}, \qquad \mathcal{L}^{(s)} = \mathcal{L}_{\text{SFT}}^{(s)} + \lambda^{(s)}\, \mathcal{L}_{\text{KL}}^{(s)}
\end{equation}
where $\epsilon > 0$ prevents the denominator from becoming too small. When $\mathcal{L}_{\text{KL}}$ is relatively small compared to $\mathcal{L}_{\text{SFT}}$, $\lambda^{(s)}$ increases to strengthen the distillation constraint; conversely, thus maintaining a dynamic balance between losses without manual tuning.

Overall, RAFT couples the two stages through the fused answer. The fused answer serves as a model-compatible target for the SFT loss and as answer context for the teacher in on-policy distillation. This design turns the teacher from a purely conservative anchor into a target-aware guide, allowing the student to acquire domain knowledge while remaining close to the behavioral patterns of the original instruction-tuned model.

\section{Experiments}

\subsection{Experimental Setup}

\paragraph{Models.}
We evaluate RAFT on three open-source instruction-tuned language models: SmolLM3-3B \citep{bakouch2025smollm3}, Llama-3.2-3B-Instruct \citep{grattafiori2024llama}, and Phi-4-mini-instruct \citep{abouelenin2025phi}. SmolLM3-3B is a compact yet capable model designed for efficient deployment, Llama-3.2-3B-Instruct is the instruction-tuned variant of Meta's Llama-3.2 series optimized for dialogue and reasoning, and Phi-4-mini-instruct is Microsoft's lightweight instruction-following model that achieves strong performance relative to its size. Evaluating across these diverse architectures allows us to verify the generalizability of our approach.

\paragraph{Domain Data.}
We construct high-quality domain-specific QA pairs by extracting question-answer pairs from the content of CCL4.0 \citep{liu2025cci40bilingualpretrainingdataset} and Common Corpus \citep{langlais2025common} using large language models, followed by a multi-model filtering process for quality assurance. The resulting dataset spans five domains: Business and Industry(B\&I), Law and Government(L\&G), Open Culture, Open Science, and Open Web, with approximately 2,000 QA pairs per domain on average.

\paragraph{Evaluation.}
We assess model performance from three complementary perspectives. For domain capability, we measure domain accuracy (\textit{D-Acc}) on the held-out test set of each domain-specific QA dataset, evaluated via majority voting among three expert models (gpt-oss-120b \citep{openai2025gptoss120bgptoss20bmodel}, DeepSeek-v3.2 \citep{deepseekai2025deepseekv32}, and Qwen3-32B \citep{qwen3technicalreport}). For general capability, we adopt two benchmarks covering different aspects: MS-Bench \citep{li2023modelscope}, a subjective question-answering benchmark also evaluated via multi-expert majority voting; and IFEval \citep{zhou2023instructionfollowingevaluationlargelanguage}, an instruction-following format evaluation. Additionally, we evaluate objective reasoning capability using MMLU \citep{hendryckstest2021,hendrycks2021ethics}, and find that all methods exhibit only minor fluctuations without significant changes (see Appendix~\ref{sec:mmlu_appendix} for details).

\paragraph{Training Details.}
We fine-tune all models for 3 epochs with a batch size of 16 and a learning rate of $5 \times 10^{-6}$. We use the AdamW optimizer with a weight decay of 0.01 and BF16 precision. In our Adaptive on-policy Distillation, we use a temperature $T=1.5$ and Top-$K=512$ for KL-based distillation. Further details are provided in Appendix~\ref{sec:T_ablation} and Section~\ref{topk}. Following Adam \citep{kingma2014adam}, the EMA decay coefficient $\beta$ is set to 0.9. All experiments are conducted on NVIDIA A100 GPUs. The primary experiments required approximately 1,000 GPU hours, with the total computational time (including preliminary trials) not exceeding 2,000 hours.

\paragraph{Baselines.}
We compare RAFT against four baselines, all using their recommended hyperparameters. \textbf{SFT} denotes standard supervised fine-tuning on the original domain data. \textbf{SDFT}~ \citep{yang2024self} is self-distillation fine-tuning, which rewrites supervision targets using the student model itself to improve compatibility. \textbf{on-policy}  \citep{lu2025onpolicydistillation} applies online on-policy distillation, directly computing KL divergence between teacher and student on student-generated trajectories. \textbf{FAPM}~ \citep{huang2025mitigating} mitigates forgetting through forgetting-aware pruning that selectively removes parameters while preserving knowledge critical for previously learned tasks.

\subsection{Main Results}

\begin{table}[!t]
\centering
\resizebox{\textwidth}{!}{%
\begin{tabular}{ll ccc ccc ccc}
\toprule
& & \multicolumn{3}{c}{SmolLM3-3B} & \multicolumn{3}{c}{Llama-3.2-3B} & \multicolumn{3}{c}{Phi-4-mini} \\
\cmidrule(lr){3-5} \cmidrule(lr){6-8} \cmidrule(lr){9-11}
Domain & Method & D-Acc & MS-Bench & IFEval & D-Acc & MS-Bench & IFEval & D-Acc & MS-Bench & IFEval \\
\midrule
\multirow{6}{*}{B\&I}
& Base Model & 13.6 & 68.0 & 31.2 & 7.4 & 46.0 & 51.2 & 9.9 & 52.5 & 55.5 \\
& SFT & 14.8 & 66.5 & 25.5 & 17.3 & 44.5 & 49.0 & 11.1 & 42.5 & 52.7 \\
\cmidrule{2-11}
& SDFT & 14.8 & 66.5 & 23.8 & 12.4 & 42.5 & 48.8 & 13.6 & 47.5 & 54.5 \\
& on-policy & 14.8 & 68.0 & 27.0 & 16.1 & \textbf{47.5} & 50.8 & \textbf{16.1} & 41.0 & 51.6 \\
& FAPM & 14.8 & 59.2 & 31.2 & 9.9 & 42.0 & 51.2 & 14.8 & \textbf{54.0} & 55.5 \\
& RAFT (Ours) & \textbf{17.3} & \textbf{71.5} & \textbf{32.0} & \textbf{19.8} & 45.5 & \textbf{51.9} & \textbf{16.1} & 44.0 & \textbf{56.8} \\
\midrule
\multirow{6}{*}{L\&G}
& Base Model & 10.3 & 68.0 & 31.2 & 7.7 & 46.0 & 51.2 & 10.3 & 52.5 & 55.5 \\
& SFT & 12.8 & 66.5 & 25.1 & 12.0 & 43.0 & 51.2 & 16.2 & 17.0 & \textbf{55.8} \\
\cmidrule{2-11}
& SDFT & 10.3 & 64.5 & 27.9 & 12.8 & 45.0 & 51.0 & 11.1 & 14.5 & 55.1 \\
& on-policy & 10.3 & 68.5 & 27.2 & 12.0 & \textbf{49.5} & 47.9 & 15.4 & 41.0 & 53.8 \\
& FAPM & 12.0 & 58.9 & \textbf{31.2} & 6.0 & 44.0 & 51.2 & 12.0 & \textbf{53.5} & 55.6 \\
& RAFT (Ours) & \textbf{18.0} & \textbf{70.0} & 30.7 & \textbf{15.4} & 43.5 & \textbf{52.5} & \textbf{18.0} & 51.5 & 53.6 \\
\midrule
\multirow{6}{*}{Sci}
& Base Model & 5.0 & 68.0 & 31.2 & 8.2 & 46.0 & 51.2 & 13.1 & 52.5 & 55.5 \\
& SFT & 8.3 & 62.0 & 23.5 & \textbf{16.4} & 31.5 & 50.1 & 18.0 & 25.0 & 52.9 \\
\cmidrule{2-11}
& SDFT & 10.0 & 58.5 & 24.4 & \textbf{16.4} & 30.0 & 48.2 & 21.3 & 30.0 & 52.5 \\
& on-policy & 11.5 & 59.5 & 27.7 & 13.1 & 32.5 & 50.1 & 21.3 & 39.5 & 53.4 \\
& FAPM & 9.8 & 59.5 & 31.2 & 11.5 & \textbf{42.5} & 51.2 & 18.0 & \textbf{53.5} & \textbf{55.3} \\
& RAFT (Ours) & \textbf{13.3} & \textbf{65.0} & \textbf{32.7} & 14.8 & 40.5 & \textbf{52.3} & \textbf{29.5} & 51.5 & 51.9 \\
\midrule
\multirow{6}{*}{Cul}
& Base Model & 9.8 & 68.0 & 31.2 & 3.3 & 46.0 & 51.2 & 8.3 & 52.5 & 55.5 \\
& SFT & 13.1 & 55.0 & 25.3 & 11.7 & 31.5 & 49.7 & \textbf{16.7} & 33.0 & 54.7 \\
\cmidrule{2-11}
& SDFT & 13.1 & 53.0 & 25.5 & 11.7 & 28.5 & 46.4 & \textbf{16.7} & 34.5 & 54.7 \\
& on-policy & 10.0 & 54.5 & 27.0 & 6.7 & 34.0 & 47.0 & 13.3 & 39.0 & 55.1 \\
& FAPM & 6.7 & 59.4 & \textbf{31.2} & 1.7 & \textbf{44.0} & \textbf{51.2} & 10.0 & \textbf{52.5} & \textbf{55.5} \\
& RAFT (Ours) & \textbf{19.7} & \textbf{65.0} & 29.9 & \textbf{13.3} & 35.5 & \textbf{51.2} & \textbf{16.7} & 37.5 & 53.2 \\
\midrule
\multirow{6}{*}{Web}
& Base Model & 14.1 & 68.0 & 31.2 & 9.7 & 46.0 & 51.2 & 16.2 & 52.5 & 55.5 \\
& SFT & 15.1 & 57.5 & 25.1 & 18.4 & 33.5 & 49.5 & 23.2 & 40.0 & 51.0 \\
\cmidrule{2-11}
& SDFT & 17.3 & 59.0 & 24.2 & 19.5 & 34.5 & 49.5 & 24.3 & 41.5 & 52.3 \\
& on-policy & 16.8 & 53.0 & 25.0 & \textbf{20.0} & 38.5 & 45.8 & 21.1 & 40.5 & 53.2 \\
& FAPM & 10.3 & 58.9 & 31.2 & 7.6 & \textbf{43.5} & \textbf{51.2} & 16.2 & \textbf{54.5} & \textbf{55.6} \\
& RAFT (Ours) & \textbf{17.8} & \textbf{62.5} & \textbf{31.8} & 19.5 & 35.5 & 50.3 & \textbf{27.6} & 31.5 & 54.2 \\
\midrule
\multirow{6}{*}{Avg}
& Base Model & 10.6 & \textbf{68.0} & 31.2 & 7.3 & \textbf{46.0} & 51.2 & 11.6 & 52.5 & \textbf{55.5} \\
& SFT & 12.8 & 61.5 & 24.9 & 15.1 & 36.8 & 49.9 & 17.1 & 31.5 & 53.4 \\
\cmidrule{2-11}
& SDFT & 13.1~{\scriptsize(+2.3)} & 60.3~{\scriptsize(-2.0)} & 25.2~{\scriptsize(+1.2)} & 14.5~{\scriptsize(-4.0)} & 36.1~{\scriptsize(-1.9)} & 48.8~{\scriptsize(-2.2)} & 17.4~{\scriptsize(+1.8)} & 33.6~{\scriptsize(+6.7)} & 53.8~{\scriptsize(+0.7)} \\
& on-policy & 12.7~{\scriptsize(-0.8)} & 60.7~{\scriptsize(-1.3)} & 26.8~{\scriptsize(+7.6)} & 13.6~{\scriptsize(-9.9)} & 40.4~{\scriptsize(+9.8)} & 48.3~{\scriptsize(-3.2)} & 17.4~{\scriptsize(+1.8)} & 40.2~{\scriptsize(+27.6)} & 53.4~{\scriptsize(+0.0)} \\
& FAPM & 10.7~{\scriptsize(-16.4)} & 59.2~{\scriptsize(-3.7)} & 31.2~{\scriptsize(+25.3)} & 7.3~{\scriptsize(-51.7)} & \textbf{43.2}~{\scriptsize(+17.4)} & 51.2~{\scriptsize(+2.6)} & 14.2~{\scriptsize(-17.0)} & \textbf{53.6}~{\scriptsize(+70.2)} & \textbf{55.5}~{\scriptsize(+3.9)} \\
& RAFT (Ours) & \textbf{17.2}~{\scriptsize(+34.0)} & \textbf{66.8}~{\scriptsize(+8.6)} & \textbf{31.4}~{\scriptsize(+26.1)} & \textbf{16.5}~{\scriptsize(+9.2)} & 40.1~{\scriptsize(+9.0)} & \textbf{51.6}~{\scriptsize(+3.5)} & \textbf{21.6}~{\scriptsize(+26.3)} & 43.2~{\scriptsize(+37.1)} & 53.9~{\scriptsize(+1.0)} \\
\bottomrule
\end{tabular}%
}
\vspace{6pt}
\caption{Main results across five domains on three backbone models. All values are percentages. \textbf{D-Acc}: domain accuracy on domain-specific datasets. Best results among fine-tuning methods are \textbf{bolded}. The \textbf{Avg} row reports the mean across all five domains for each metric. In the Avg row, the relative improvement over SFT is shown in parentheses.}
\label{tab:main}
\end{table}


As shown in Table~\ref{tab:main}, RAFT achieves the strongest average trade-off across the three backbone models and five domains. It does not dominate every individual model-domain setting, but it consistently improves the averaged D-Acc over standard SFT and recovers a substantial portion of the SFT-induced degradation on MS-Bench and IFEval. Averaging the backbone-level relative improvements, RAFT improves D-Acc by 23.2\%, MS-Bench by 18.2\%, and IFEval by 10.2\% over SFT. We also report retention rates relative to the base model in Appendix~\ref{app:retention}. RAFT improves average MS-Bench retention from 76.8\% to 89.2\% and IFEval retention from 91.2\% to 99.5\% compared with SFT, supporting that its gains come with better preservation of general capabilities.


Among the baselines, each exhibits clear limitations. FAPM maintains IFEval performance on SmolLM3-3B (31.2\%, on par with the base model), but achieves only marginal domain gains (D-Acc of 10.7\%, even below SFT's 12.8\%), indicating that pruning-based regularization alone is insufficient for effective domain adaptation. SDFT and on-policy distillation show mixed performance: while they occasionally outperform SFT on specific metrics (e.g., SDFT on Phi-4-mini achieves +1.8\% D-Acc and +6.7\% MS-Bench), such improvements are not consistent across models; for instance, both methods degrade D-Acc on Llama-3.2-3B-Instruct. These contrasting failure modes reveal that addressing only one dimension, training stage constraints or data refinement is insufficient. RAFT selects the most model-compatible samples to refine training data and adds distillation constraints to the objective. This approach achieves a stronger average trade-off between domain learning and forgetting prevention.



\subsection{Ablation Studies}
We conduct controlled ablation studies to validate the key components of RAFT. Specifically, fused supervision is shown to provide higher-quality and more informative training signals compared to original supervision. Adaptive on-policy distillation plays a key role in improving general capability preservation during domain adaptation. Furthermore, EMA-based adaptive loss balancing is essential for dynamically resolving the trade-off between domain learning and general capability retention that fixed coefficients cannot address.

\subsubsection{Data Ablation: Effect of Fused Supervision}

To evaluate the contribution of our data fusion strategy, we compare standard SFT using original domain data versus using fused data. Table~\ref{tab:ablation} presents the results on SmolLM3-3B across three representative domains.

\begin{table}[t]                                                                                                                                                          
  \centering                                                                                                                                                                
  \begin{tabularx}{0.8\columnwidth}{l l *{3}{>{\centering\arraybackslash}X}}                                                                                                   
  \toprule                                                                                                                                                                  
  Domain & Method & D-Acc & MS-Bench & IFEval \\                                                                                                                            
  \midrule                                                                                                                                                                  
  \multirow{3}{*}{L\&G}                                                                                                                                                     
  & Original Data + SFT & 12.8 & 66.5 & 25.1 \\                                                                                                                             
  & Fused Data + SFT & 13.7 & 69.5 & 28.3 \\                                                                                                                                
  & RAFT (Ours) & \textbf{18.0} & \textbf{70.0} & \textbf{30.7} \\                                                                                                          
  \cmidrule{2-5}                                                                                                                                                            
  \multirow{3}{*}{Sci}                                                                                                                                                      
  & Original Data + SFT & 8.3 & 62.0 & 23.5 \\                                                                                                                              
  & Fused Data + SFT & 10.0 & 64.0 & 27.2 \\                                                                                                                                
  & RAFT (Ours) & \textbf{13.3} & \textbf{65.0} & \textbf{32.7} \\                                                                                                          
  \cmidrule{2-5}                                                                                                                                                            
  \multirow{3}{*}{Cul}                                                                                                                                                      
  & Original Data + SFT & 13.1 & 55.0 & 25.3 \\                                                                                                                             
  & Fused Data + SFT & 14.8 & 58.5 & 25.3 \\                                                                                                                                
  & RAFT (Ours) & \textbf{19.7} & \textbf{65.0} & \textbf{29.9} \\                
  \bottomrule     
  \end{tabularx}                              
  \vspace{6pt}                                   
  \caption{Ablation study on data fusion and adaptive on-policy distillation on SmolLM3-3B across three domains. All values are percentages. Best results among all settings
   are \textbf{bolded}.}                                                                                                                                                    
  \label{tab:ablation}
  \end{table}


Fused data consistently outperforms original data across all three domains. The domain accuracy improves by 7.0\% on Law \& Government, 20.5\% on Open Science, and 13.0\% on Open Culture, while MS-Bench also improves by 4.5\%, 3.2\%, and 6.4\% respectively. IFEval improves by 12.7\% on Law \& Government and 15.7\% on Open Science, and remains unchanged on Open Culture. These results demonstrate that fusing the original and self-distilled answers via a strong teacher produces more informative and better-aligned supervision signals, which help reduce distribution shift and mitigate catastrophic forgetting.

\subsubsection{Training Objective Ablation: Effect of Adaptive On-Policy Distillation}

To evaluate Adaptive On-Policy Distillation, we compare standard SFT and the full RAFT method. Both settings use the same fused data, ensuring that any performance difference can be attributed to the adaptive on-policy distillation objective. Table~\ref{tab:ablation} presents the results on SmolLM3-3B.


The adaptive on-policy distillation objective provides substantial gains over fused-data SFT alone. Domain accuracy improves by 31.4\% on Law \& Government, 33.0\% on Open Science, and 33.1\% on Open Culture. MS-Bench improves modestly by 0.7\%, 1.6\%, and 11.1\% respectively, while IFEval improves by 8.5\%, 20.2\%, and 18.2\%. These results confirm that aligning the student with the teacher on its own generated trajectories, combined with EMA-based adaptive weighting, effectively constrains the training process and further mitigates forgetting while enhancing domain performance. The consistent improvements across all domains demonstrate that the training objective is complementary to the data fusion strategy.





\subsubsection{Impact of Adaptive Loss Balancing}
\begin{figure}[t]
  \centering       
  \begin{subfigure}[t]{0.48\textwidth}                                                                                             
      \centering                                                                                                                   
      \includegraphics[width=\textwidth]{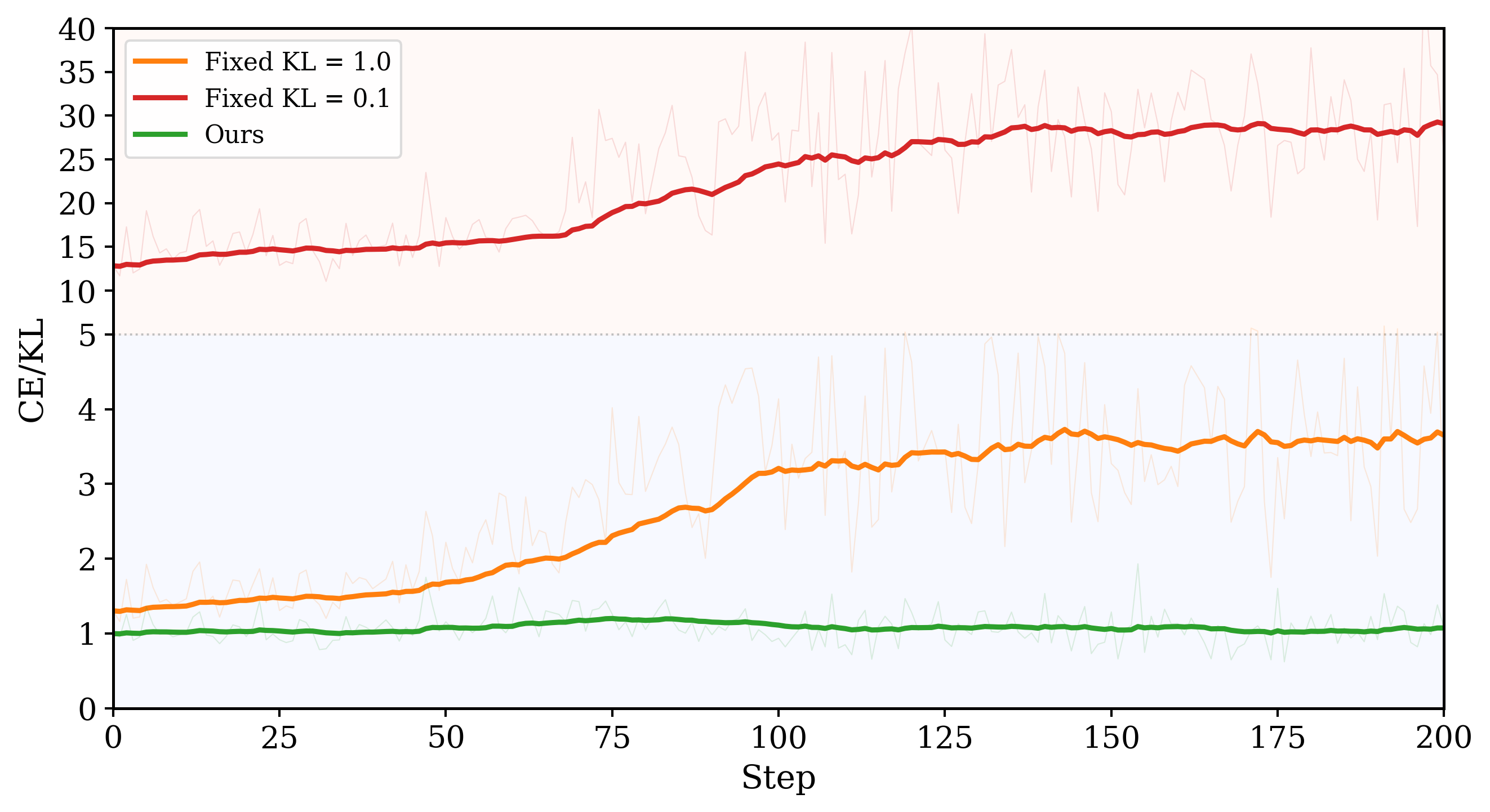}                                                  
      \caption{CE/KL loss ratio curves under different balancing strategies on SmolLM3-3B in the Open Culture domain. ``Fixed''    
  denotes using a constant KL coefficient ($\lambda=0.1$ or $\lambda=1.0$), while ``Ours'' denotes the EMA-based adaptive scheme.}
      \label{fig:ema_loss}
  \end{subfigure}
  \hfill
  \begin{subfigure}[t]{0.48\textwidth}
      \centering
      \includegraphics[width=\textwidth]{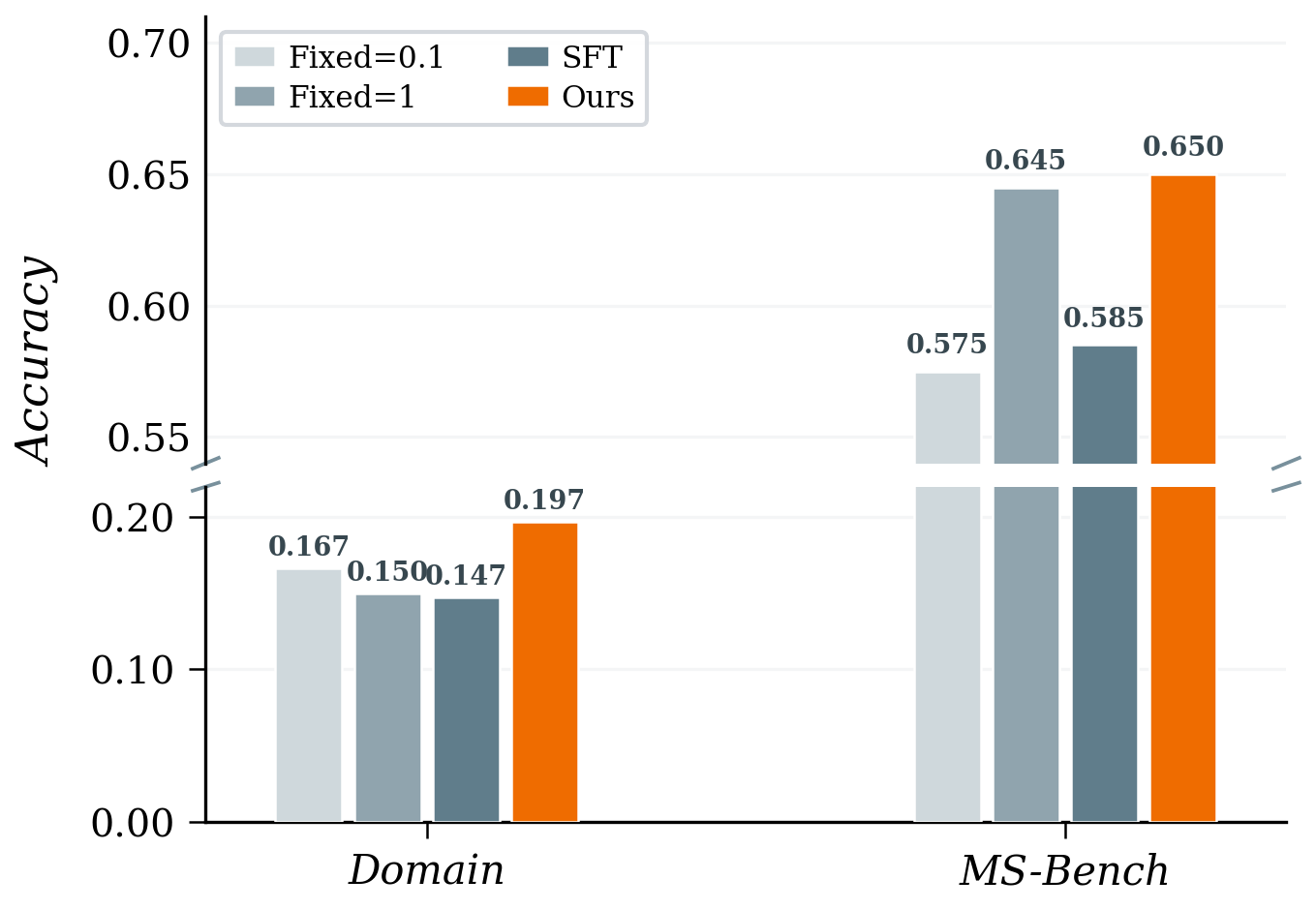}
      \caption{Domain accuracy (Open Culture datasets) and general capability (MS-Bench) of SmolLM3-3B under different balancing
  strategies. Standard SFT, $\lambda=0.1$, $\lambda=1.0$, and EMA-based adaptive scheme (Ours) are compared.}
      \label{fig:ema_accuracy}
  \end{subfigure}
  \caption{Effect of different balancing strategies on SmolLM3-3B in the Open Culture domain.}
  \label{fig:ema_ablation}
  \end{figure}
A key design choice in RAFT is the EMA-based adaptive balancing mechanism that dynamically adjusts the relative weight $\lambda^{(s)}$ between $\mathcal{L}_{\text{SFT}}$ and $\mathcal{L}_{\text{KL}}$ during training. We compare our adaptive scheme against fixed KL coefficients $\lambda \in \{0.1, 1.0\}$ and standard SFT (no distillation) on SmolLM3-3B in the Open Culture domain.

\paragraph{Training stability.} Figure~\ref{fig:ema_loss} shows the CE/KL loss ratio curves under different balancing strategies. With both fixed coefficients ($\lambda=0.1$ and $\lambda=1.0$), the ratio fluctuates sharply,
  indicating that the relative weight between CE and KL losses is highly unstable. In contrast, the EMA-based adaptive scheme maintains a stable ratio throughout training, 
  indicating that the CE loss and KL loss are well-balanced and the optimization proceeds smoothly.
\paragraph{Domain and general trade-off.} Figure~\ref{fig:ema_accuracy} presents the domain accuracy and general capability under different balancing strategies. With a fixed coefficient, there exists an inherent trade-off: increasing $\lambda$ strengthens regularization and better preserves general capability (MS-Bench rises from 57.5\% at $\lambda=0.1$ to 64.5\% at $\lambda=1.0$), but simultaneously suppresses domain adaptation (D-Acc drops from 16.7\% to 15.0\%). Standard SFT without distillation achieves a D-Acc of 14.7\% and MS-Bench of 58.5\%, suffering from both limited domain adaptation and general capability degradation. No single fixed coefficient can simultaneously optimize both objectives. The adaptive scheme resolves this dilemma. It dynamically adjusts the weight based on the relative magnitudes of the two losses. It balances the distillation constraint by approximately equalizing the scalar magnitudes of the two loss terms, ensuring $\lambda \mathcal{L}_{KL}$ remains comparable to $\mathcal{L}_{SFT}$. This yields the best domain accuracy of 19.7\% and the best MS-Bench score of 65.0\%, outperforming all fixed-coefficient baselines on domain adaptation and achieving a favorable balance on general capability preservation.

These results demonstrate that EMA-based adaptive balancing is crucial for RAFT: it eliminates the need for manual tuning of the distillation coefficient, stabilizes training, and enables a dynamic trade-off that fixed coefficients cannot achieve.

\section{Conclusions}

We propose RAFT, a two-stage framework for domain-specific fine-tuning with alleviated forgetting. RAFT couples model-compatible data refinement with answer-conditioned trajectory-level preservation: the offline stage constructs higher-quality fused supervision, while the adaptive on-policy distillation stage regularizes student-generated trajectories using the original instruction-tuned model as a target-aware teacher. Across three backbone models and five domains, RAFT achieves a stronger averaged trade-off than standard SFT and the evaluated forgetting-mitigation baselines. Averaged across all settings, RAFT improves D-Acc by 23.2\% over standard SFT, while improving MS-Bench and IFEval by 18.2\% and 10.2\%, respectively. These results suggest that coupling compatible supervision with trajectory-level preservation provides a practical recipe for domain adaptation with better general capability retention.



\section{Limitations}
This work has two primary limitations. Firstly, due to computational constraints, our evaluation of RAFT is focused on small-scale language models. While RAFT demonstrates consistent improvements across multiple architectures, its scalability to larger models warrants further investigation. Secondly, the diversity of training data and domains is relatively constrained. Although we evaluated RAFT across five distinct domains, its performance in specialized tasks such as code generation, multilingual contexts, or significantly larger training corpora remains to be explored. 
{
\small
\bibliographystyle{plainnat}
\bibliography{references}


 }

\newpage
\appendix

\section{Model Similarity Analysis}
\label{sec:model_similarity_appendix}

\begin{figure}[H]
\centering
\includegraphics[width=0.7\textwidth]{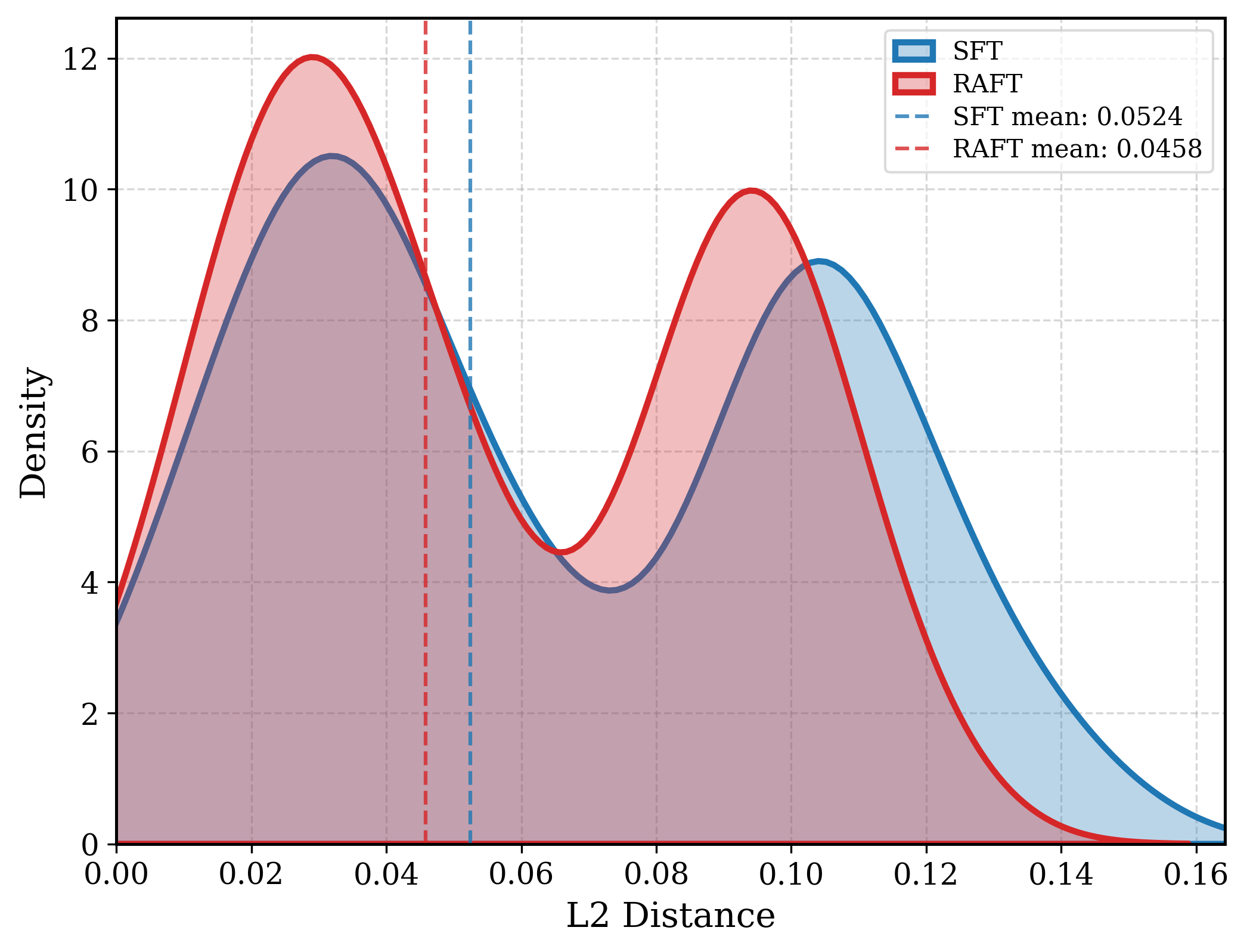}
\caption{Distribution of relative $\ell_2$ weight distances from the pre-trained SmolLM3-3B model. Blue: standard SFT (mean = 0.0524). Red: RAFT (mean = 0.0458). A smaller distance indicates less deviation from the pre-trained model and thus less catastrophic forgetting. RAFT produces weights that are consistently closer to the original model across all layers.}
\label{fig:weight_distance}
\end{figure}

To further quantify the degree of catastrophic forgetting at the parameter level, we measure the $\ell_2$ distance between the fine-tuned model weights and the original instruction-tuned checkpoint. A smaller distance indicates that the fine-tuned model remains closer to the original model, suggesting that less pre-trained knowledge has been overwritten during fine-tuning. Figure~\ref{fig:weight_distance} shows the distribution of relative $\ell_2$ distances for standard SFT and RAFT on SmolLM3-3B.

RAFT achieves a mean $\ell_2$ distance of 0.0458 compared to 0.0524 for standard SFT, a 12.6\% relative reduction. This provides parameter-level evidence that RAFT produces more conservative updates than standard SFT. Although weight distance isn't the only factor in forgetting, this smaller deviation aligns with the better performance seen on MS-Bench and IFEval. The reduced weight deviation is consistent with RAFT's better preservation of general capabilities observed in the main results, further corroborating the connection between parameter-level conservation and functional-level forgetting prevention.

\section{Comparison with Related Fine-Tuning Strategies}
\label{app:method-comparison}
Table~\ref{tab:method-comparison} summarizes how RAFT differs from related forgetting-mitigation strategies. RAFT is designed to couple model-compatible supervision with trajectory-level preservation. In contrast, data-only methods improve the reference answers but do not constrain student-generated trajectories, while optimization-only methods regularize trajectories or parameters without explicitly improving domain supervision.
\begin{table}[t]
  \centering
  \begin{tabularx}{\columnwidth}{l *{4}{>{\centering\arraybackslash}X}}
  \toprule
  Method & Data refinement & On-policy trajectory & Answer-conditioned teacher & Adaptive balancing \\
  \midrule
  SFT & -- & -- & -- & -- \\
  SDFT & \checkmark & -- & -- & -- \\
  Replay-based methods & -- & -- & -- & mixing ratio \\
  On-policy distillation & -- & \checkmark & -- & fixed / manual \\
  FAPM & -- & -- & -- & -- \\
  \textbf{RAFT (Ours)} & \textbf{\checkmark} & \textbf{\checkmark} & \textbf{\checkmark} & \textbf{\checkmark} \\
  \bottomrule
  \end{tabularx}
  \vspace{6pt}
  \caption{Conceptual comparison between RAFT and related domain fine-tuning strategies.}
  \label{tab:method-comparison}
\end{table}

\section{Response Fusion Prompt Template}
\label{app:fusion_prompt}

To enhance the quality of model-generated responses, we use a response fusion strategy. It combines an original answer and a model-regenerated answer into a single higher-quality response. The fusion is guided by a carefully designed prompt. It integrates complementary information from both responses while removing redundancy and preserving factual accuracy and logical coherence.

The complete prompt template used for response fusion is presented below:

\begin{lstlisting}
You are a professional knowledge assistant. Below is a user question along with two different responses (Response 1 is the original answer, and Response 2 is a regenerated answer based on the original). Please integrate the strengths and knowledge from both, and generate a higher-quality new response.

Requirements:
- Synthesize information from both responses to make the content more complete, accurate, and clearly expressed;
- Avoid repetition and redundancy;
- Preserve factual correctness, professionalism, and logical coherence;
- Output only the best fused answer you can produce;
- Strictly output the answer only, without any additional content.

---

Question:
{query}

Response 1 (Original):
{answer1}

Response 2 (Model-generated):
{answer2}

Fused final response:
\end{lstlisting}

\section{Quality Scoring Prompt}
  \label{appendix:Quality_Scoring}        

In the fusion stage, we use a single expert to generate the fused output. We then employ multiple judge models to independently evaluate its consistency with the original answer. The detailed prompt used for quality scoring is as follows:

\subsection{Prompt Template}

\begin{lstlisting}                                                 
You are a strict, fair, and hallucination-aware automatic scoring assistant.

Task Description:
- You will receive a Question, a Reference Answer, and a Model Answer.
- Your task is to determine whether the Model Answer correctly covers the core content of the Reference Answer, and to check if the model exhibits hallucination (i.e., fabricating content that is irrelevant, contradictory, or invalid with respect to the question or reference answer).
- If the model contains additional information, as long as it is semantically consistent with the question and reference answer without conflict, no points should be deducted.
- If there is fabrication of facts, logical errors, or deviation from the topic, it should be considered hallucination and result in a lower score.

Scoring Criteria:
- 1.0: Completely correct. The model answer accurately expresses the core information of the reference answer with no hallucination.
- 0.8 - 0.9: Mostly correct. The expression is highly consistent with the reference answer with only minor wording differences, and no obvious hallucination.
- 0.6 - 0.7: Partially correct. Although some core content is mentioned, there are omissions or minor inaccuracies, with occasional questionable extensions.
- 0.3 - 0.5: Weakly related. Missing key content, or containing obviously irrelevant/invalid descriptions (hallucination).
- 0.0 - 0.2: Incorrect or irrelevant. Fails to express the main content of the reference answer, or contains severe hallucination, contradictions, or fabricated information.

---

Question:
{user_question}

Reference Answer:
{original_answer}

Model Answer:
{fused_answer}

Please only output a score from 0 to 1
\end{lstlisting}

\subsection{Aggregation Strategy}                                                       
For each fused output, we collect scores from $M$ independent judge models and compute the average score $\bar{c}$. We perform multiple rounds of fusion and scoring, and 
  select the result with the highest average score as the final output. Moreover, to reduce computational overhead, we adopt a threshold of $0.8$: if the average score
  $\bar{c} \geq 0.8$, the fused output is considered to be of sufficiently high quality and is adopted immediately without further generation rounds, thereby reducing      
  unnecessary resource consumption. This multi-model scoring mechanism ensures robust quality selection while tolerating individual judge variability.

\section{Parametric Analysis}
Considering the impact of different parameter values on RAFT, here we have conducted a detailed analysis of the relevant parameters.
\subsection{Analysis of Cosine Similarity Threshold $\tau$}
\label{appendix:cos_threshold}

In the semantic filtering stage of fusion supervision construction (Section~\ref{sec:fusion}), the cosine similarity threshold $\tau$ controls the trade-off between adopting the student model's self-rewritten answer and falling back to the original ground truth answer. A lower $\tau$ accepts more rewrites, which better adapts the expression style to the student model but risks introducing semantic drift; a higher $\tau$ is more conservative, preserving factual accuracy but potentially missing opportunities to reduce distribution mismatch. We analyze the effect of $\tau \in \{0, 0.3, 0.5, 0.8, 1.0\}$ on SmolLM3-3B in the Open Culture domain, where both domain accuracy and general capability are evaluated.

\begin{figure}[H]
\centering
\includegraphics[width=0.85\textwidth]{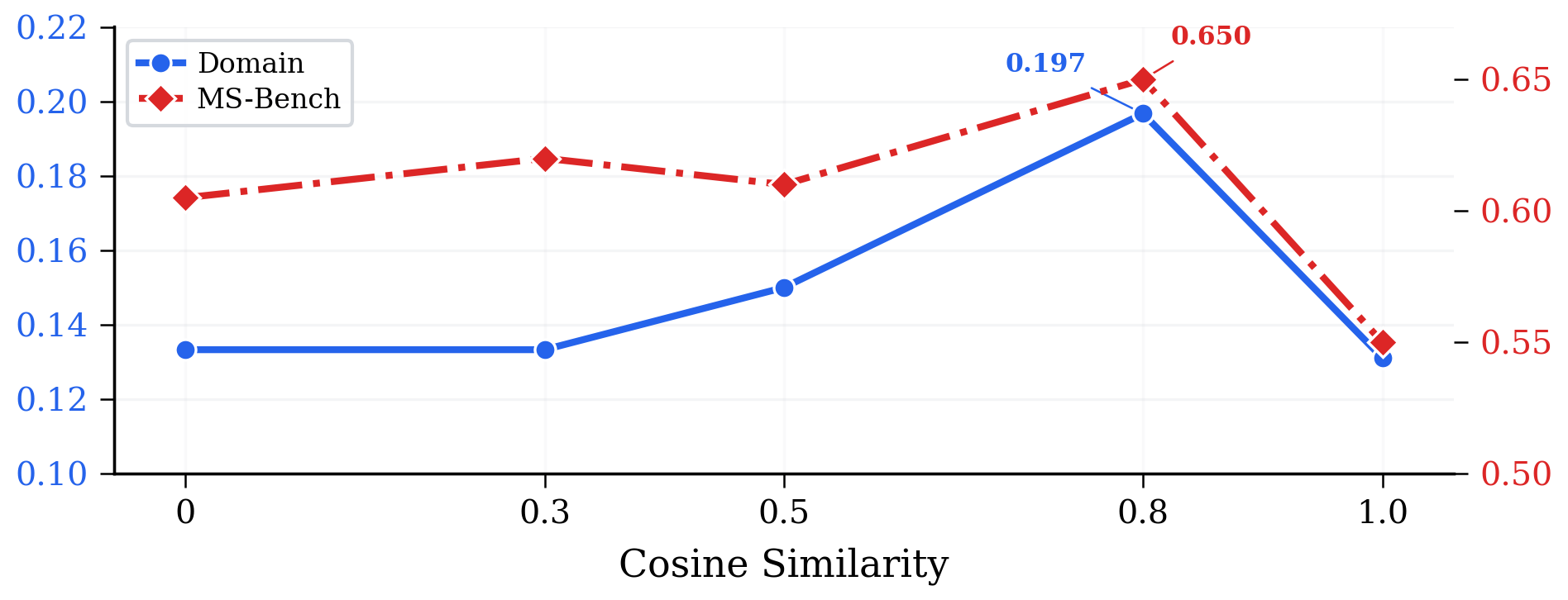}
\caption{Effect of cosine similarity threshold $\tau$ on domain accuracy (D-Acc) on Open Culture and general capability (MS-Bench) on SmolLM3-3B. Both metrics peak at $\tau=0.8$, indicating that a moderately strict filtering threshold best balances semantic fidelity and distribution adaptation.}
\label{fig:cos_threshold}
\end{figure}

As shown in Figure~\ref{fig:cos_threshold}, both domain accuracy and general capability exhibit a non-monotonic relationship with $\tau$. At the two extremes, the results are suboptimal for different reasons. When $\tau=0$, all rewrites are accepted without filtering, allowing semantic drift to corrupt the training data---this yields a D-Acc of 13.3\% and an MS-Bench of 60.5\%. When $\tau=1.0$, the threshold is so strict that virtually no rewrite can satisfy the condition ($s_i \ge 1.0$ requires exact identity),this similarly yields a D-Acc of 13.1\% and an MS-Bench of 55.0\%.

At intermediate thresholds, both metrics improve as $\tau$ increases, reaching their joint optimum at $\tau=0.8$ with a D-Acc of 19.7\% and an MS-Bench of 65.0\%. This result suggests that a moderately strict threshold effectively filters out semantically drifted rewrites while retaining those that successfully adapt the expression style to the student model's distribution. The improved training data quality, in turn, benefits both domain adaptation and general capability preservation. The consistent improvement across both metrics at $\tau=0.8$ confirms that semantic filtering plays a critical role in the fusion supervision pipeline: it is not merely a safeguard against errors, but an active mechanism that enhances the overall quality of the training data. We therefore adopt $\tau=0.8$ as the default throughout all experiments.

\subsection{Temperature Ablation}
\label{sec:T_ablation}

\begin{figure}[t]
\centering
\includegraphics[width=0.85\textwidth]{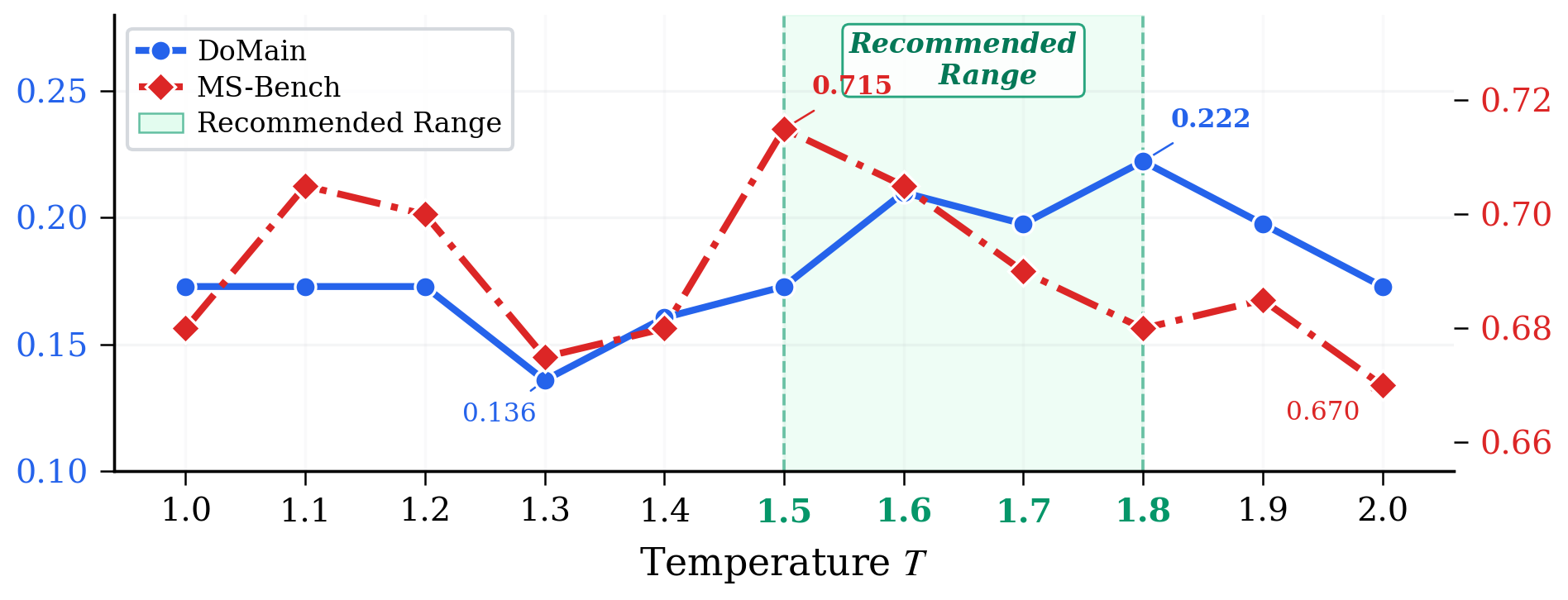}
\caption{Effect of temperature $T$ on D-Acc and MS-Bench on SmolLM3-3B (Business \& Industry). The shaded region indicates the recommended range $T \in [1.5, 1.8]$.}
\label{fig:T_ablation}
\end{figure}

The temperature $T$ controls the smoothness of the teacher distribution in KL divergence computation. We ablate $T$ from 1.0 to 2.0 on SmolLM3-3B in the Business \& Industry domain. As shown in Figure~\ref{fig:T_ablation}, D-Acc and MS-Bench exhibit different sensitivities to $T$: D-Acc rises steadily with $T$ (from 17.3\% at $T=1.0$ to 22.2\% at $T=1.8$), while MS-Bench peaks at $T=1.5$ (71.5\%) and declines beyond. Excessively low $T$ yields sharp teacher distributions with limited supervision, while excessively high $T$ over-smooths and dilutes discriminative information. We select $T=1.5$ as the default, achieving the best MS-Bench (71.5\%) with competitive D-Acc (17.3\%).

\subsection{Detailed Analysis of Top-$K$ Selection}
\label{topk}

In our distillation framework, the Top-$K$ parameter restricts the KL divergence computation to the $K$ tokens with the highest teacher probabilities, serving a dual purpose: reducing the computational cost of full-vocabulary KL and focusing distillation on the most informative tokens. We analyze the effect of Top-$K$ from two perspectives---probability coverage versus computational cost, and downstream task performance---to justify our choice of $K=512$.

\begin{figure}[t]
\centering
\includegraphics[width=0.85\textwidth]{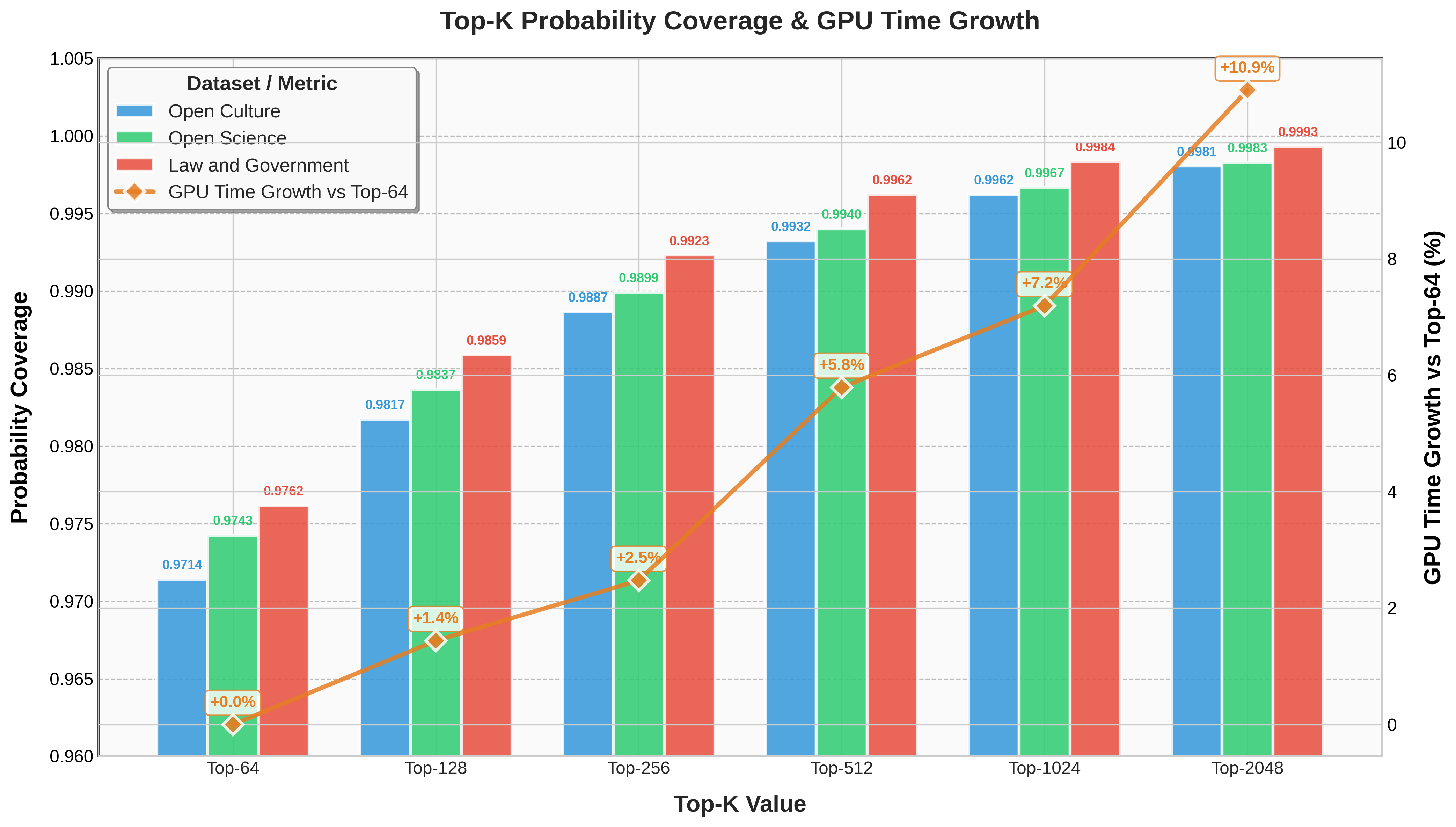}
\caption{Probability coverage and relative GPU time growth under different Top-$K$ values. The bars show the cumulative probability coverage of the teacher distribution captured by the Top-$K$ tokens; the line shows the relative GPU time increase compared to $K=32$. As $K$ increases beyond 512, the marginal coverage gain diminishes rapidly while computational cost continues to grow.}
\label{fig:topk_coverage}
\end{figure}

\paragraph{Probability Coverage and Computational Cost.}
Figure~\ref{fig:topk_coverage} presents the probability coverage of the teacher distribution and the relative GPU time overhead under different Top-$K$ values. The teacher's probability distribution is highly concentrated: Top-256 already covers 99.2\% of the probability mass, and Top-512 achieves 99.6\% coverage. Beyond 512, the marginal coverage gain diminishes rapidly (99.8\% at 1024, 99.9\% at 2048), while GPU time continues to grow. This indicates that the vast majority of the teacher's knowledge is captured within the top 512 tokens, and increasing $K$ further yields negligible information gain at disproportionately higher computational cost.

\begin{figure}[t]
\centering
\includegraphics[width=0.85\textwidth]{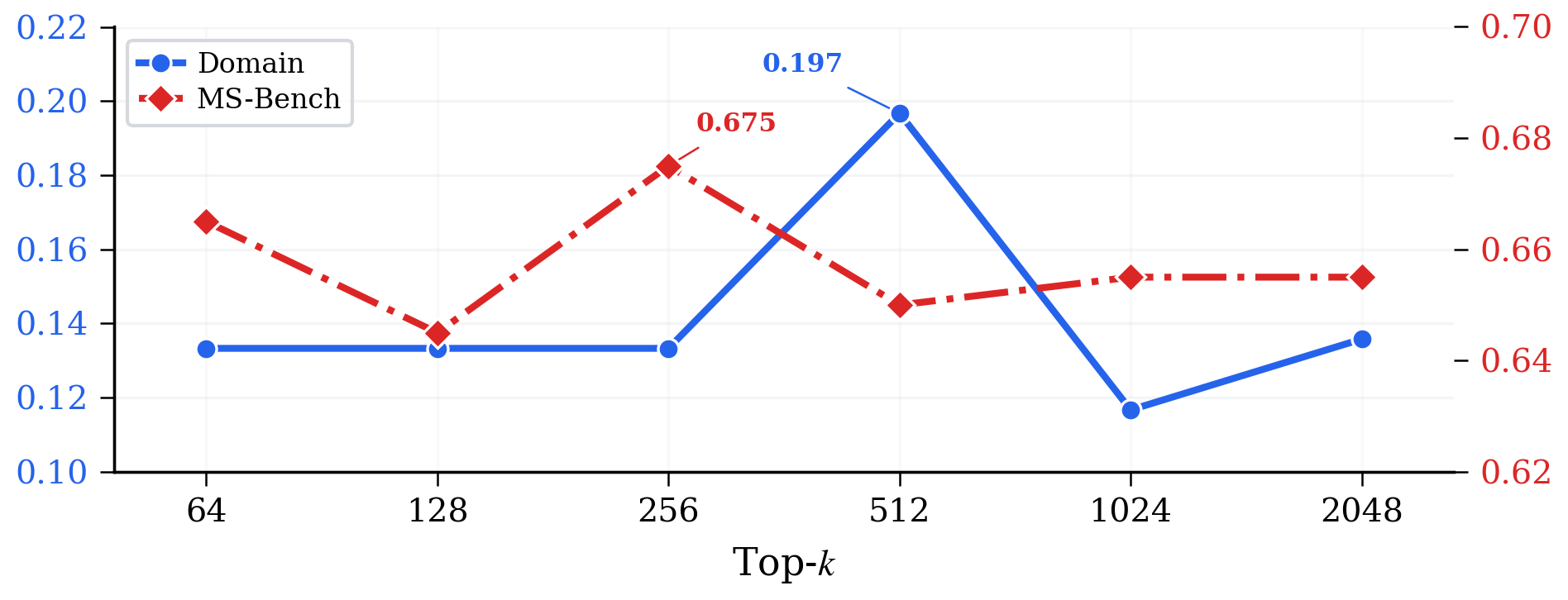}
\caption{Domain accuracy (D-Acc) on Open Culture and general capability (MS-Bench) under different Top-$K$ values on SmolLM3-3B. D-Acc peaks at $K=512$ while MS-Bench remains relatively stable across all $K$ values, indicating that the choice of $K$ primarily affects domain adaptation quality rather than general capability preservation.}
\label{fig:topk_accuracy}
\end{figure}

\paragraph{Effect of Top-$K$ on Task Performance.}
To examine how Top-$K$ affects both domain adaptation and general capability preservation, we conduct experiments on SmolLM3-3B in the Open Culture domain with $K \in \{64, 128, 256, 512, 1024, 2048\}$. Figure~\ref{fig:topk_accuracy} shows the domain accuracy (D-Acc) and general capability (MS-Bench) under different $K$ values.

As shown in Figure~\ref{fig:topk_accuracy}, the two metrics exhibit markedly different sensitivities to $K$. MS-Bench remains relatively stable across all $K$ values, fluctuating within a narrow range of 64.5\%--67.5\%, which suggests that general capability preservation is largely determined by the presence of the distillation constraint itself rather than the specific number of tokens involved. In contrast, domain accuracy shows a clear non-monotonic trend: D-Acc stays at 13.3\% for $K \le 256$, rises sharply to 19.7\% at $K=512$, and then drops to 11.7\% at $K=1024$ before partially recovering to 13.6\% at $K=2048$.

We attribute this non-monotonic behavior to the trade-off between information sufficiency and noise. When $K$ is too small (e.g., $K \le 256$), the restricted token set fails to capture enough of the teacher's distribution to provide effective guidance for domain adaptation, resulting in suboptimal D-Acc. At $K=512$, the distilled knowledge is both sufficiently comprehensive and sufficiently focused, yielding the best domain performance. When $K$ becomes too large (e.g., $K=1024$), low-probability tokens that carry little discriminative information are included in the KL computation, which can introduce noise and dilute the distillation signal, leading to degraded D-Acc. The partial recovery at $K=2048$ suggests that with a sufficiently large $K$, the full-vocabulary KL approaches a stable regime where the noise effect is averaged out, though at significantly higher computational cost.

\paragraph{Summary.}
Combining both analyses, $K=512$ represents the optimal operating point: it captures 99.6\% of the teacher's probability mass with acceptable computational overhead, and achieves the highest domain accuracy while maintaining stable general capability. We therefore adopt $K=512$ as the default throughout all experiments.

\section{MMLU Results}
\label{sec:mmlu_appendix}

\begin{figure}[t]
\centering
\includegraphics[width=0.85\textwidth]{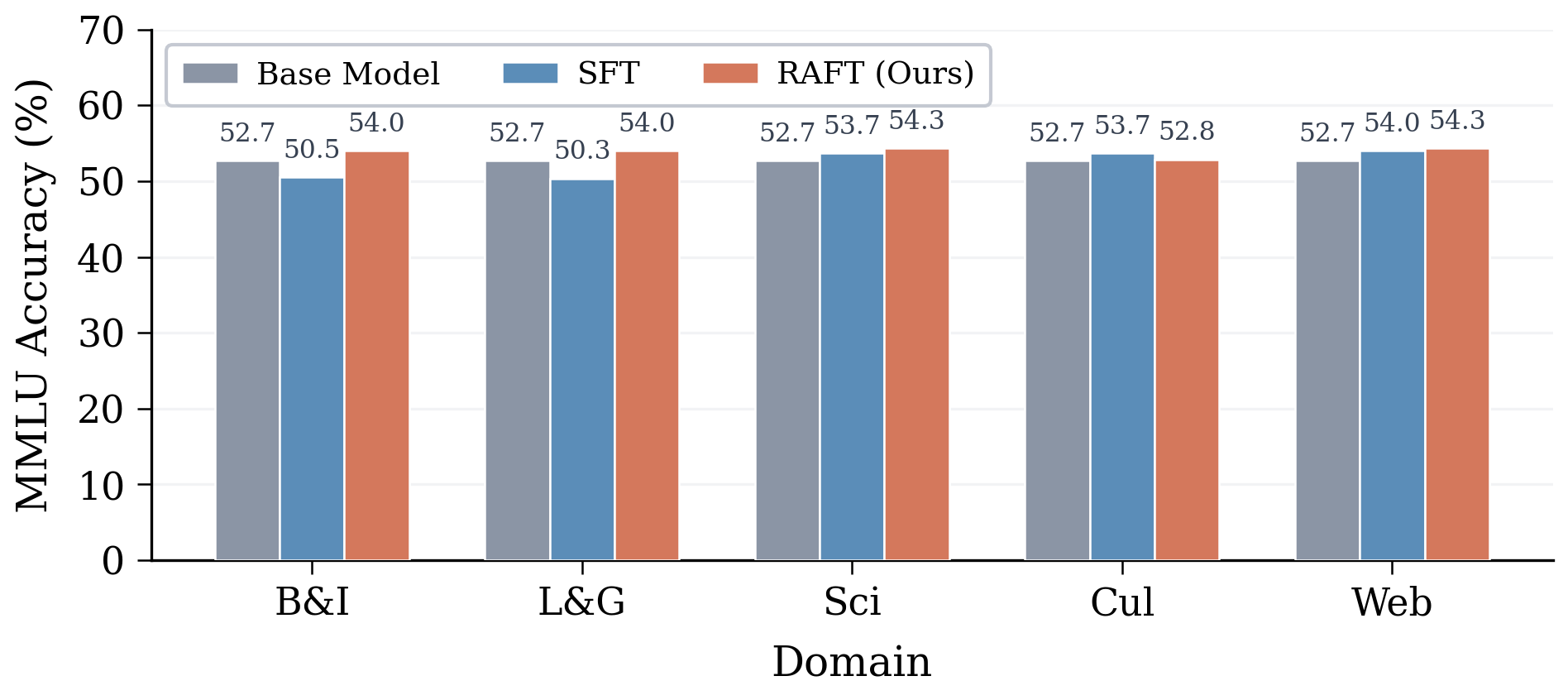}
\caption{MMLU accuracy of the base model, SFT, and RAFT across five domains on SmolLM3-3B. All methods show comparable MMLU performance, indicating that RAFT preserves objective reasoning capability during domain adaptation.}
\label{fig:mmlu}
\end{figure}

We further examine the impact of domain adaptation on objective reasoning through the MMLU benchmark on SmolLM3-3B. As shown in Figure~\ref{fig:mmlu}, we compare the MMLU accuracy of the model, standard SFT, and RAFT across five domains. All three methods exhibit comparable MMLU performance, with values ranging from 50.3\% to 54.3\% across domains. The differences among the base model, SFT, and RAFT are marginal (within 2 percentage points in each domain), indicating that domain adaptation via RAFT does not compromise objective reasoning capability. RAFT achieves the highest or near-highest scores on four out of five domains (54.0\% on B\&I, 54.0\% on L\&G, 54.3\% on Sci, and 54.3\% on Web), closely matching the base model's 52.7\% and SFT's 50.3\%--54.0\%. These results demonstrate that RAFT effectively preserves objective reasoning while achieving substantial domain adaptation gains.

\section{General Capability Retention Analysis}
\label{app:retention}
\begin{table}[t]
\centering
\small
\setlength{\tabcolsep}{4.5pt}

\begin{tabular}{llrrrrrr}
\toprule
Backbone & Metric & Base & SFT & RAFT & SFT Ret. & RAFT Ret. & $\Delta$ Ret. \\
\midrule
SmolLM3-3B & MS-Bench & 68.0 & 61.5 & 66.8 & 90.4\% & 98.2\% & +7.8 \\
SmolLM3-3B & IFEval   & 31.2 & 24.9 & 31.4 & 79.8\% & 100.6\% & +20.8 \\
\midrule
Llama-3.2-3B & MS-Bench & 46.0 & 36.8 & 40.1 & 80.0\% & 87.2\% & +7.2 \\
Llama-3.2-3B & IFEval   & 51.2 & 49.9 & 51.6 & 97.5\% & 100.8\% & +3.3 \\
\midrule
Phi-4-mini & MS-Bench & 52.5 & 31.5 & 43.2 & 60.0\% & 82.3\% & +22.3 \\
Phi-4-mini & IFEval   & 55.5 & 53.4 & 53.9 & 96.2\% & 97.1\% & +0.9 \\
\midrule
Average & MS-Bench & -- & -- & -- & 76.8\% & 89.2\% & +12.4 \\
Average & IFEval   & -- & -- & -- & 91.2\% & 99.5\% & +8.4 \\
\bottomrule
\end{tabular}
\caption{
General capability retention relative to the base model. Retention is computed as
$\mathrm{Retention} = \mathrm{Score}_{\mathrm{method}} / \mathrm{Score}_{\mathrm{base}} \times 100\%$.
The values are derived from the averaged results in Table~\ref{tab:main}.
}
\label{tab:general-retention}
\end{table}

To make the forgetting effect more explicit, we additionally report the general capability retention rate relative to the corresponding base model. For a general benchmark score $S$, the retention rate is defined as
\[
\mathrm{Retention}(\mathrm{method}) =
\frac{S_{\mathrm{method}}}{S_{\mathrm{base}}} \times 100\%.
\]
We compute this value only for the general capability benchmarks, namely MS-Bench and IFEval, because D-Acc measures domain adaptation rather than retention of the original model's general ability.
Table~\ref{tab:general-retention} shows that RAFT improves general capability retention over standard SFT on both MS-Bench and IFEval across all three backbone models. On MS-Bench, the average retention increases from 76.8\% under SFT to 89.2\% under RAFT. On IFEval, the average retention increases from 91.2\% to 99.5\%. These results provide a complementary view of the main results: RAFT improves domain accuracy over SFT while better retaining the base model's general capabilities on the evaluated benchmarks.

\section{Teacher Model Concatenation Template}
\label{appendix:template}

In the Answer-Conditioned On-Policy Distillation, the teacher model conditions on the fused answer $\hat{y}_i$ as additional context when computing logits on the student's generated trajectory. This appendix provides the detailed concatenation template.

\subsection{Template Structure}

The teacher model's input sequence is constructed by concatenating the following components in order:

\begin{lstlisting}
[Prefix] + [Fused Answer] + [Middle] + [User Prompt] + [Suffix] + [Student Trajectory]
\end{lstlisting}

\noindent where:
\begin{itemize}
    \item \textbf{Prefix}: ``Background reference knowledge:''
    \item \textbf{Fused Answer}: $\hat{y}_i$, the high-quality fused answer from Section 3.2
    \item \textbf{Middle}: ``\textbackslash nAnswer the question based on the reference knowledge:\textbackslash n''
    \item \textbf{User Prompt}: $x_i$, the original user instruction
    \item \textbf{Suffix}: ``\textbackslash nAnswer:''
    \item \textbf{Student Trajectory}: $\mathbf{z}_i$, the token sequence generated by the student model
\end{itemize}

\subsection{Complete Template Example}

The complete input template for the teacher model is as follows:

\begin{lstlisting}
Background reference knowledge: {fused_answer}
Answer the question based on the reference knowledge:
{user_prompt}
Answer: {student_trajectory}
\end{lstlisting}

\noindent In contrast, the student model only conditions on:

\begin{lstlisting}
{user_prompt}{student_trajectory}
\end{lstlisting}

\subsection{Design Rationale}

This asymmetric context design serves two purposes:

\begin{enumerate}
    \item \textbf{Informed guidance}: By conditioning on $\hat{y}_i$, the teacher model can provide more informative soft targets at each position of the student's trajectory, as it ``knows'' the reference answer while evaluating the student's generation.

    \item \textbf{Distribution alignment}: The teacher's logits on the student's trajectory reflect how a model with access to ground-truth knowledge would assign probabilities, guiding the student toward better generalization while preserving its own generation style.
\end{enumerate}

\subsection{Implementation Details}

In practice, the template tokens are encoded without special tokens and concatenated as tensor sequences:

\begin{lstlisting}
teacher_input_ids = concat([
    encode("Background reference knowledge:"),
    fused_answer_ids,
    encode("\nAnswer the question based on "
           "the reference knowledge:\n"),
    prompt_ids,
    encode("\nAnswer:"),
    trajectory_ids
], dim=1)
\end{lstlisting}

\noindent The logits corresponding to the trajectory positions are then extracted for distillation loss computation, ensuring alignment between teacher and student predictions on the same token positions.

\end{document}